# Multi-Stage CNN Architecture for Face Mask Detection

Amit Chavda, Jason Dsouza, Sumeet Badgujar, Ankit Damani

iPing Data Labs LLP, Mumbai

amit.chavda@iping.in

jason.dsouza@iping.in

sumeet.badgujar@iping.in

ankit.damani@iping.in

**Abstract:**
The end of 2019 witnessed the outbreak of Coronavirus Disease 2019 (COVID-19), which has continued to be the cause of plight for millions of lives and businesses even in 2020. As the world recovers from the pandemic and plans to return to a state of normalcy, there is a wave of anxiety among all individuals, especially those who intend to resume in-person activity. Studies have proved that wearing a face mask significantly reduces the risk of viral transmission as well as provides a sense of protection. However, it is not feasible to manually track the implementation of this policy. Technology holds the key here. We introduce a Deep Learning based system that can detect instances where face masks are not used properly. Our system consists of a dual-stage Convolutional Neural Network (CNN) architecture capable of detecting masked and unmasked faces and can be integrated with pre-installed CCTV cameras. This will help track safety violations, promote the use of face masks, and ensure a safe working environment.

**Keywords:**
Computer Vision; Object Detection; Object Tracking; COVID-19; Face Masks; Safety Improvement

## 1. Introduction:

Rapid advancements in the fields of Science and Technology have led us to a stage where we are capable of achieving feats that seemed improbable a few decades ago. Technologies in fields like Machine Learning and Artificial Intelligence have made our lives easier and provide solutions to several complex problems in various areas. Modern Computer Vision algorithms are approaching human-level performance in visual perception tasks. From image classification to video analytics, Computer Vision has proven to be a revolutionary aspect of modern technology. In a world battling against the Novel Coronavirus Disease (COVID-19) pandemic, technology has been a lifesaver. With the aid of technology, 'work from home' has substituted our normal work routines and has become a part of our daily lives. However, for some sectors, it is impossible to adapt to this new norm.

As the pandemic slowly settles and such sectors become eager to resume in-person work, individuals are still skeptical of getting back to the office. 65% of employees are now anxious about returning to the office (Woods, 2020). Multiple studies have shown that the use of face masks reduces the risk of viral transmission as well as provides a sense of protection (Howard et al., 2020; Verma et al., 2020). However, it is infeasible to manually enforce such a policy on large premises and track any violations. Computer Vision provides a better alternative to this. Using a combination of image classification, object detection, object tracking, and video analysis, we developed a robust system that can detect the presence and absence of face masks in images as well as videos.

In this paper, we propose a two-stage CNN architecture, where the first stage detects human faces, while the second stage uses a lightweight image classifier to classify the faces detected in the first stage as either 'Mask' or 'No Mask' faces and draws bounding boxes around them along with the detected class name. This algorithm was further extended to videos as well. The detected faces are then tracked between frames using an object tracking algorithm, which makes the detections robust to the noise due to motion blur. This system can then be integrated with an image or video capturing device like a CCTV camera, to track safety violations, promote the use of face masks, and ensure a safe working environment.

## 2. Methods:
### 2.1. Related Work:
*2.1.1. Traditional Object Detection:*
The problem of detecting multiple masked and unmasked faces in images can be solved by a traditional object detection model. The process of object detection mainly involves localizing the objects in images and classifying them (in case of multiple objects). Traditional algorithms like Haar Cascade (Viola and Jones, 2001) and HOG (Dalal and Triggs, 2005) have proved to be effective for such tasks, but these algorithms are heavily based on Feature Engineering. In the era of Deep learning, it is possible to train Neural Networks that outperform these algorithms, and do not need any extra Feature Engineering.

*2.1.2. Convolutional Neural Networks:*
Convolutional Neural Networks (CNNs) (LeCun et al., 1998) is a key aspect in modern Computer Vision tasks like pattern object detection, image classification, pattern recognition tasks, etc. A CNN uses convolution kernels to convolve with the original images or feature maps to extract higher-level features, thus resulting in a very powerful tool for Computer Vision tasks.

*2.1.3. Modern Object Detection Algorithms:*
CNN based object detection algorithms can be classified into 2 categories: Multi-Stage Detectors and Single-Stage Detectors.

Multi-Stage Detectors: In a multi-stage detector, the process of detection is split into multiple steps. A two-stage detector like RCNN (Girshick et al., 2014) first estimates and proposes a set of regions of interest using selective search. The CNN feature vectors are then extracted from each region independently. Multiple algorithms based on Regional Proposal Network like Fast RCNN (Girshick, 2015) and Faster RCNN (Ren et al., 2015) have achieved higher accuracy and better results than most single stage detectors.

Single-Stage Detectors: A single-stage detector performs detections in one step, directly over a dense sampling of possible locations. These algorithms skip the region proposal stage used in multi-stage detectors and are thus considered to be generally faster, at the cost of some loss of accuracy. One of the most popular single-stage algorithms, You Only Look Once (YOLO) (Redmon et al., 2016), was introduced in 2015 and achieved close to real-time performance. Single Shot Detector (SSD) (Liu et al., 2016) is another popular algorithm used for object detection, which gives excellent results. RetinaNet (Lin et al., 2017b), one of the best detectors, is based on Feature Pyramid Networks (Lin et al., 2017a), and uses focal loss.

*2.1.4. Face Mask Detection:*
As the world began implementing precautionary measures against the Coronavirus, numerous implementations of Face Mask Detection systems came forth.

(Ejaz et al., 2019) have performed facial recognition on masked and unmasked faces using Principal Component Analysis (PCA). However, the recognition accuracy drops to less than 70% when the recognized face is masked.

(Qin and Li, 2020) introduced a method to identify face mask wearing conditions. They divided the facemask wearing conditions into three categories: correct face mask wearing, incorrect face mask wearing, and no face mask wearing. Their system takes an image, detects and crops faces, and then uses SRCNet (Dong et al., 2016) to perform image super-resolution and classify them.

The work by (Nieto-Rodríguez et al., 2015) presented a method that detects the presence or absence of a medical mask. The primary objective of this approach was to trigger an alert only for medical staff who do not wear a surgical mask, by minimizing as many false-positive face detections as possible, without missing any medical mask detections.

(Loey et al., 2021) proposed a model that consists of two components. The first component performs uses ResNet50 (He et al., 2016) for feature extraction. The next component is a facemask classifier, based on an ensemble of classical Machine Learning algorithms. The authors evaluated their system and estimated that Deep Transfer Learning approaches would achieve better results since the building, comparing, and selecting the best model among a set of classical Machine Learning models is a time-consuming process.

## 2.2. Proposed Methodology:

We propose a two-stage architecture for detecting masked and unmasked faces and localizing them.

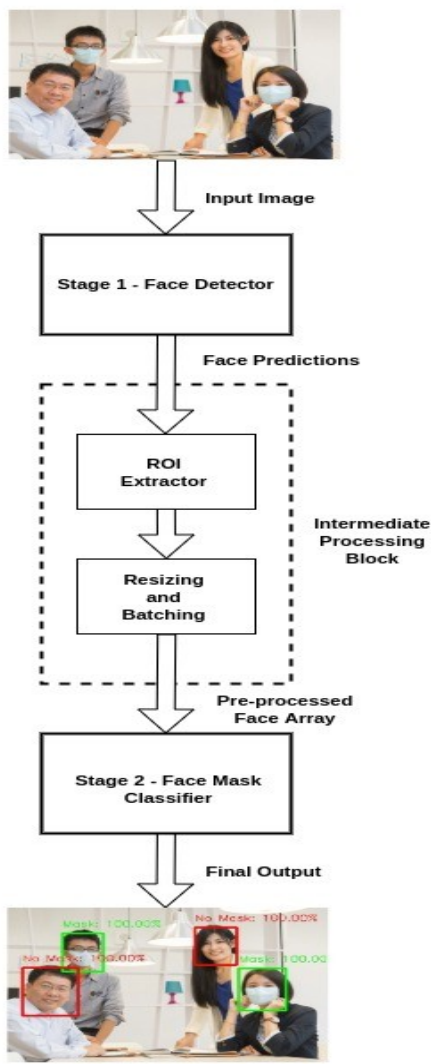

**Fig. 1: System Architecture**

### 2.2.1. Architecture Overview:

Fig. 1 represents our proposed system architecture (input image taken from the dataset by (Larxel, 2020)). It consists of two major stages. The first stage of our architecture includes a Face Detector, which localizes multiple faces in images of varying sizes and detects faces even in overlapping scenarios. The detected faces (regions of interest) extracted from this stage are then batched together and passed to the second stage of our architecture, which is a CNN based Face Mask Classifier. The results from the second stage are decoded and the final output is the image with all the faces in the image correctly detected and classified as either masked or unmasked faces.

### 2.2.2. Stage 1 - Face Detector:

A face detector acts as the first stage of our system. A raw RGB image is passed as the input to this stage. The face detector extracts and outputs all the faces detected in the image with their bounding box coordinates. The process of detecting faces accurately is very important for our architecture. Training a highly accurate face detector needs a lot of labeled data, time, and compute resources. For these reasons, we selected a pre-trained model trained on a large dataset for easy generalization and stability in detection. Three different pre-trained models were tested for this stage:

**Dlib** (Sharma et al., 2016) - The Dlib Deep Learning face detector offers significantly better performance than its precursor, the Dlib HOG based face detector.

**MTCNN** (Zhang, K. et al, 2016) - It uses a cascade architecture with three stages of CNN for detecting and localizing faces and facial keypoints.

**RetinaFace** (Deng et al., 2020) - It is a single-stage design with pixel-wise localization that uses a multi-task learning strategy to simultaneously predict face box, face score, and facial keypoints.

The detection process is challenging for the model used in this stage, as it needs to detect human faces that could also be covered with masks. We selected RetinaFace as our Stage 1 model, based on our experimentation and comparative analysis, covered in section 3.2.

*2.2.3. Intermediate Processing Block:*

This block carries out the processing of the detected faces and batches them together for classification, which is carried out by Stage 2. The detector from Stage 1 outputs the bounding boxes for the faces. Stage 2 requires the entire head of the person to accurately classify the faces as masked or unmasked. The first step involves expanding the bounding boxes in height and width by 20%, which covers the required Region of Interest (ROI) with minimal overlap with other faces in most situations. The second step involves cropping out the expanded bounding boxes from the image to extract the ROI for each detected face. The extracted faces are resized and normalized as required by Stage 2. Furthermore, all the faces are batched together for batch inference.

*2.2.4. Stage 2 - Face Mask Classifier:*

The second stage of our system is a face mask classifier. This stage takes the processed ROI from the Intermediate Processing Block and classifies it as either Mask or No Mask. A CNN based classifier for this stage was trained, based on three different image classification models: MobileNetV2 (Sandler et al., 2018), DenseNet121 (Huang et al., 2017), NASNet (Zoph et al., 2018). These models have a lightweight architecture that offers high performance with low latency, which is suitable for video analysis. The output of this stage is an image (or video frame) with localized faces, classified as masked or unmasked.

*2.2.5. Dataset:*

The three face mask classifier models were trained on our dataset. The dataset images for masked and unmasked faces were collected from image datasets available in the public domain, along with some data scraped from the Internet. Masked images were obtained from the Real-world Masked Face Recognition Dataset (RMFRD) (Wang, Z. et al., 2020) and Face Mask Detection dataset by Larxel on Kaggle (Larxel, 2020).

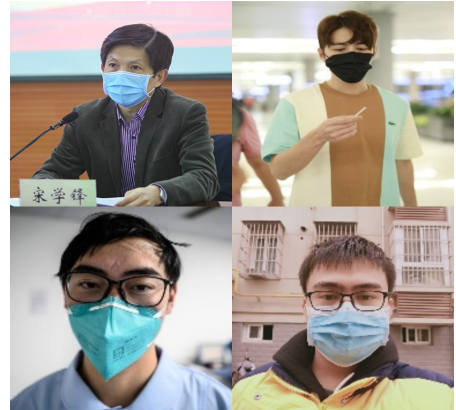

**Fig. 2: RMFRD Masked Images**

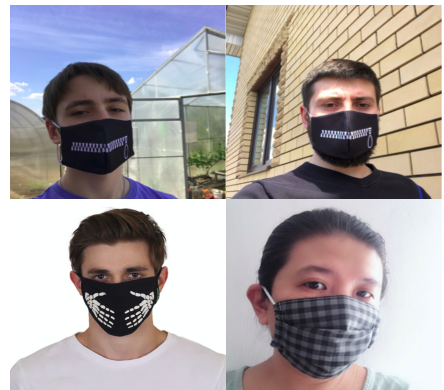

**Fig. 3: Larxel (Kaggle) Masked Images**

RMFRD images were biased towards Asian faces. Thus, masked images from the Larxel (Kaggle) were added to the dataset to eliminate this bias. RMFRD contains images for unmasked faces as well. However, as mentioned before, they were heavily biased towards Asian faces. Hence, we decided not to use these images. The Flickr-Faces-HQ (FFHQ) dataset introduced by (Karras et al., 2019) was used for unmasked images.

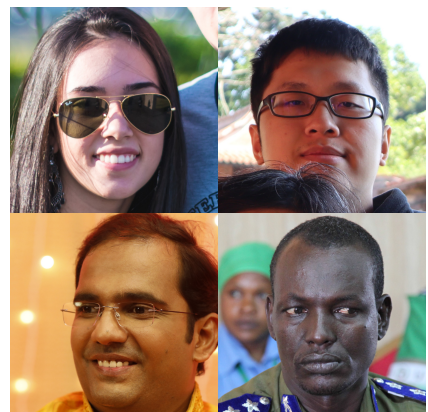

**Fig. 4: Flickr Faces HQ Non-Masked Images**

Our dataset also includes images of improperly worn face masks or hands covering the face, which get classified as non-masked faces.

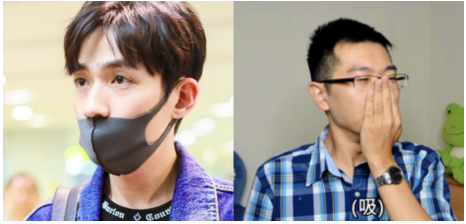

**Fig. 5: Images with improperly worn face masks or hands covering the face**

The collected RAW data was passed through Stage 1 (Face Detector) and the Intermediate Processing Block of the architecture. This process was carried out to ensure that the distribution and nature of training data for Stage 2 match the expected input for Stage 2 during the final deployment.

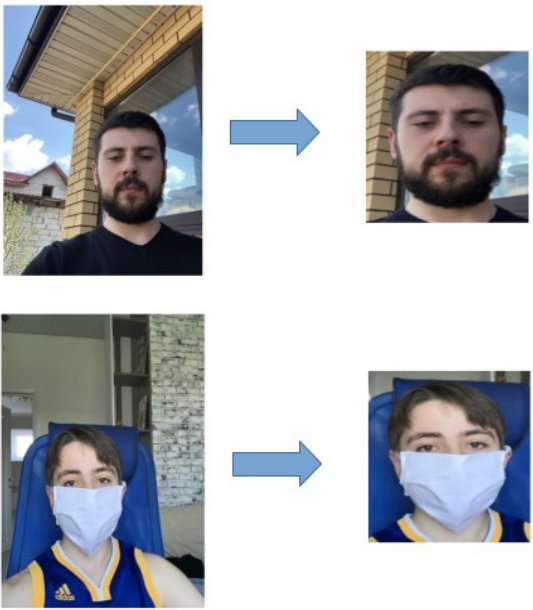

**Fig. 6: RAW Images converted to their respective ROIs using the Intermediate Processing Block**

The final dataset has **7855** images, divided into two classes:

| Class Name | Description | No. of images |
|---|---|---|
| Mask | Faces with masks correctly used | 3440 |
| No_Mask | Faces with no masks or masks incorrectly used | 4415 |

**Table 1: Face Mask Classifier Dataset**

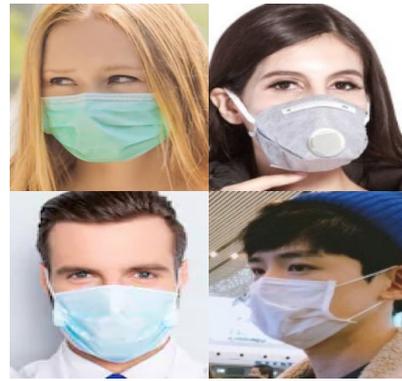

**Fig. 7: Final Dataset Images for 'Mask' Class**

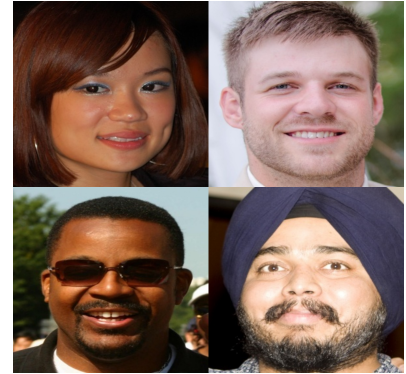

**Fig. 8: Final Dataset Images for 'No_Mask' Class**

*2.2.6. Face Mask Classifier Model Training:*
For the second stage, three CNN classifiers were trained for classifying images as masked or unmasked. The models were trained using the Keras (Chollet et al., 2015) framework. Pre-trained ImageNet (Deng et al., 2009) weights were used as a starting point for these models, instead of Glorot Uniform Initializer (Glorot and Bengio, 2010). The dataset was split into train, validation, and test sets in a ratio of 80:10:10. Data augmentation was performed using the ImageDataGenerator class in Keras. The input image size was set as 224 x 224. We selected an initial learning rate of 0.001. Besides this, the training process included checkpointing the weights for best loss, reducing the learning rate on plateau, and early stopping. Each model was trained for 50 epochs and the weights from the epoch with the lowest validation loss were selected. Based on a comparative analysis of performance, covered in sections 3.1 and 3.3, the weights trained using the NASNetMobile architecture were chosen as our final trained weights.

## 3. Experimental Analysis:
(All images used in section 3 are either self-obtained or belong to the dataset by (Larxel, 2020))

### 3.1. Face Mask Classifier Training Statistics:

| Model | Training | | Validation | | Test Acc. (%) |
|---|---|---|---|---|---|
| | Acc. (%) | Loss | Acc. (%) | Loss | |
| NASNet Mobile | **99.82** | **0.0012** | 99.45 | **0.0181** | 99.23 |
| Dense Net121 | 99.49 | 0.0157 | **98.73** | 0.0312 | **99.49** |
| Mobile NetV2 | 99.42 | 0.0181 | 99.36 | 0.0297 | 99.23 |

**Table 2: Face Mask Classifier Training Statistics**

Based on Table 2, we can say that all three models have achieved very good statistics. The NASNetMobile model has overall slightly better numbers than the other two models.

| Model | Precision | Recall | F1-Score |
|---|---|---|---|
| NASNetMobile | 98.28 | **100** | 99.13 |
| DenseNet121 | **99.70** | 99.12 | **99.40** |
| MobileNetV2 | 99.12 | 99.12 | 99.12 |

**Table 3: Face Mask Classifier Performance Metrics**

Table 3 shows that DenseNet121 has the best F1-Score. However, the other models are not significantly behind. Thus, there was a need to measure other aspects of performance comparison like inference speed and model size, to select the final Face Mask Classifier Model.

### 3.2. Face Detector Comparison:
We tested three pre-trained models for face detection in Stage 1: Dlib DNN, MTCNN, RetinaFace. The average inference times for each of the models were calculated, based on a set of masked and unmasked images. As observed in Table 4, the RetinaFace model performs the best.

| Model | Average Inference Time per image (in seconds), for varying image resolutions | | |
|---|---|---|---|
| | 480p | 720p | 1080p |
| Dlib DNN | 3.622 | 8.206 | 14.654 |
| MTCNN | 0.575 | 0.684 | 1.260 |
| RetinaFace | 0.095 | 0.113 | 0.196 |

**Table 4: Face Detector Inference Speed**

It was observed that all three models show good results on images taken from a very short distance, having no more than two people in the image. However, it was noticed that as the number of people in the images increases, the performance of Dlib becomes subpar. Dlib also struggles to detect masked or covered faces.

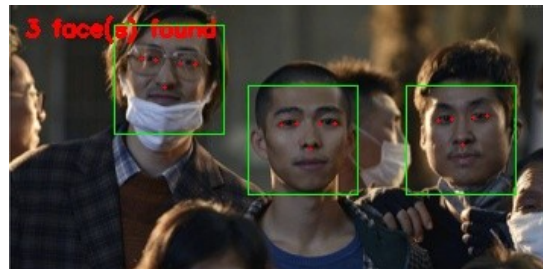

**Fig. 9 (a). Dlib good detection on normal faces**

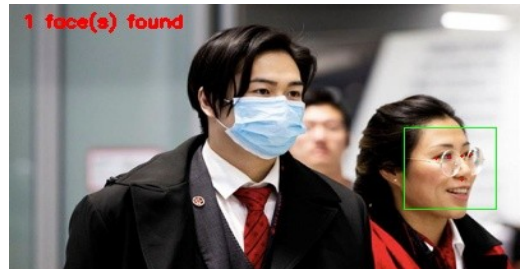

**Fig. 9 (b). Dlib poor detection on faces covered by face masks**

MTCNN and RetinaFace perform better than Dlib and can detect multiple faces in images. Both of them can detect masked or covered faces as well. MTCNN has very high accuracy when detecting faces from the front view, but its accuracy heavily drops when detecting faces from the side view.

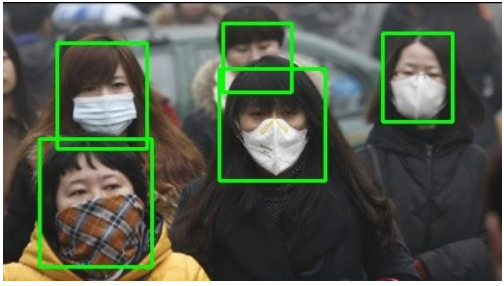

**Fig. 10 (a). MTCNN good detection on covered faces**

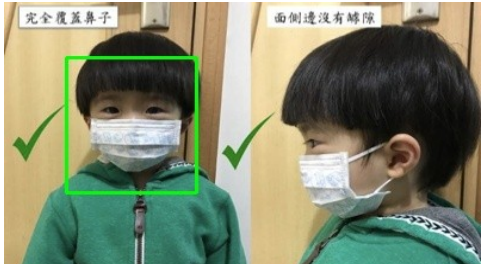

**Fig. 10 (b). MTCNN poor detection on faces in side view**

On the other hand, RetinaFace can detect side view faces with good accuracy as well. Compared to MTCNN, RetinaFace significantly decreases the failure rate from 26.31% to 9.37% (the NME threshold at 10%) (Deng et al., 2020, Page 6).

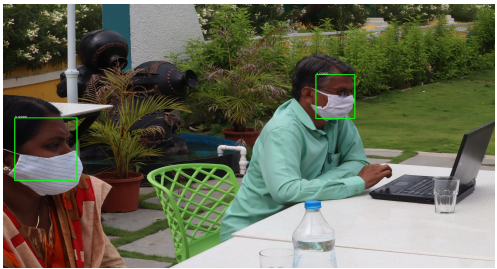

**Fig. 11 (a). RetinaFace good detection on covered faces**

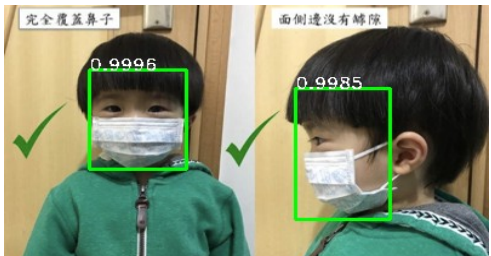

**Fig. 11 (b). RetinaFace good detection on faces in side view**

Therefore, we decided to use RetinaFace as our Face Detector for Stage 1.

### 3.3. Face Mask Classifier Comparison:

| Model | Average Inference Time for a 720p resolution image (in seconds) | No. of network parameters (in millions) |
|---|---|---|
| NASNetMobile | 0.295 | 4.88 |
| DenseNet121 | 0.353 | 8.52 |
| MobileNetV2 | 0.118 | 4.07 |

**Table 5: Face Mask Classifier Inference Speed and Model Size Comparison**

NASNetMobile and DenseNet121 give better results than MobileNetV2 and are almost on par with each other. From the observations in Table 5, it is evident that NASNet performs much faster than DenseNet121. Furthermore, the model size of NASNet is lighter than DenseNet121 (due to a lesser number of parameters). This leads to faster loading of the model during inference. Due to these factors, NASNetMobile is much more suited for real-time applications as compared to DenseNet121. Therefore, NASNetMobile was selected as our final model for the Face Mask Classifier.

### 3.4. Final Results:

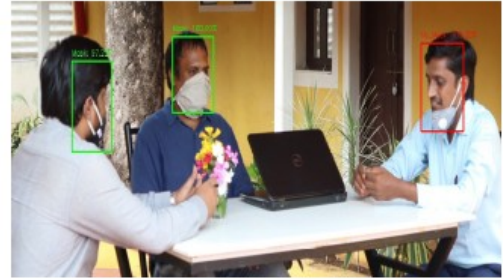

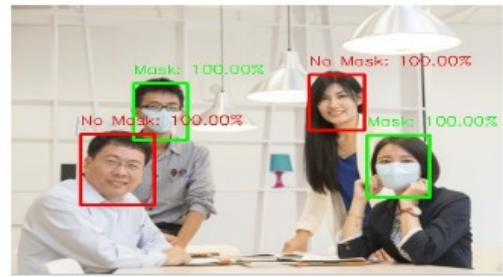

**Fig. 12: Final Results**

Combining all the components of our architecture, we thus get a highly accurate and robust Face Mask Detection System. RetinaFace was selected as our Face Detector in Stage 1, while the NASNetMobile based model was selected as our Face Mask Classifier in Stage 2. The resultant system exhibits high performance and has the capability to detect face masks in images with multiple faces over a wide range of angles.

### 3.5. Video Analysis:

Until now, we have seen that our system shows high performance over images, overcoming most of the issues commonly faced in object detection in images. For real-world scenarios, it is beneficial to extend such a detection system to work over video feeds as well.

Videos have their own set of challenges like motion blur, dynamic focus, transitioning between frames, etc. In order to ensure that the detections remain stable and to avoid jitter between frames, we used the process of Object Tracking.

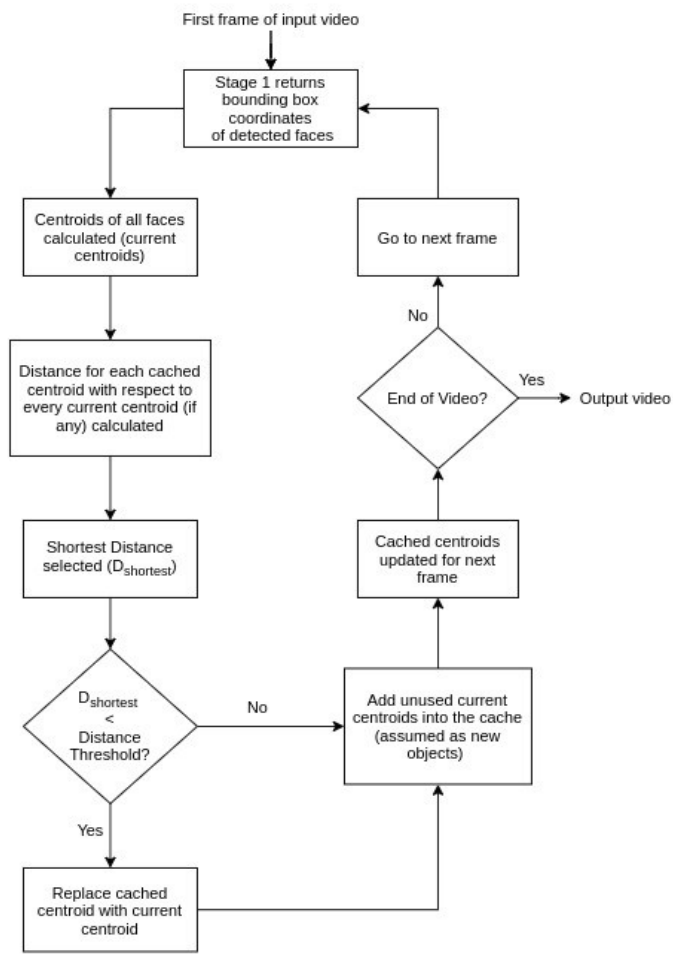

**Fig. 13: Integration of Centroid Tracking in Our System**

We used a modified version of Centroid Tracking, inspired by (Nascimento et al., 1999), in order to track the detected faces between consecutive frames. This makes our detection algorithm robust to the noise and the motion blur in video streams, where the algorithm could fail to detect some objects.

The detected face ROIs in a given frame are tracked over a predefined number of frames so that the ROI coordinates for the faces are stored even if the detector fails to detect the object during the transition between frames.

We selected five frames as the threshold in 30 FPS video streams for discarding the cached centroids, which gave good results with the least false positive face detections in video streams. After using this method, there was a significant improvement in face mask detection in video streams. The following results show the difference in detection with and without Centroid Tracking:

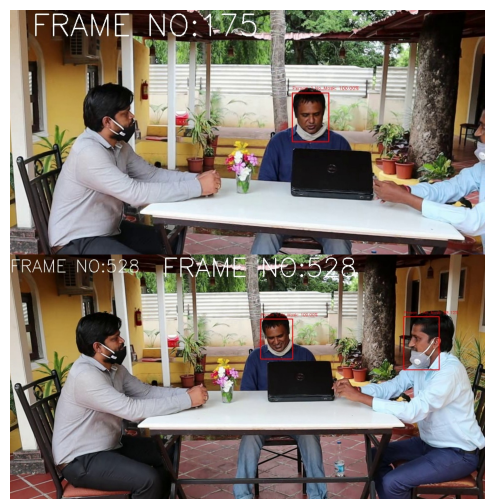

**Fig. 14 (a). Without Centroid Tracking**

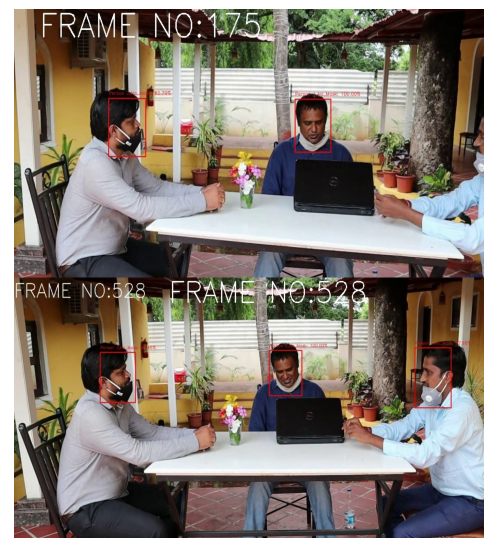

**Fig. 14 (b). Same Frames with Centroid Tracking**

## 4. Discussion:

### 4.1. Conclusions:

In this paper, a two-stage Face Mask Detector was presented. The first stage uses a pre-trained RetinaFace model for robust face detection, after comparing its performance with Dlib and MTCNN. An unbiased dataset of masked and unmasked faces was created. The second stage involved training three different lightweight Face Mask Classifier models on the created dataset and based on performance, the NASNetMobile based model was selected for classifying faces as masked or non-masked. Furthermore, Centroid Tracking was added to our algorithm, which helped improve its performance on video streams. In times of the COVID-19 pandemic, with the world looking to return to normalcy and people resuming in-person work, this system can be easily deployed for automated monitoring of the use of face masks at workplaces, which will help make them safer.

### 4.2. Future Scope:

There are a number of aspects we plan to work on shortly:

- Currently, the model gives 5 FPS inference speed on a CPU. In the future, we plan to improve this up to 15 FPS, making our solution deployable for CCTV cameras, without the need of a GPU.
- The use of Machine Learning in the field of mobile deployment is rising rapidly. Hence, we plan to port our models to their respective TensorFlow Lite versions.
- Our architecture can be made compatible with TensorFlow RunTime (TFRT), which will increase the inference performance on edge devices and make our models efficient on multithreading CPUs.
- Stage 1 and Stage 2 models can be easily replaced with improved models in the future, that would give better accuracy and lower latency.